\documentclass[9pt, conference]{IEEEtran}
\IEEEoverridecommandlockouts
\usepackage{cite}
\usepackage{amsmath,amssymb,amsfonts}
\usepackage{algorithmic}
\usepackage{booktabs}
\usepackage{graphicx}
\usepackage{textcomp}
\usepackage{xcolor}
\usepackage{multirow}
\usepackage{algorithm}
\usepackage{multirow}
\usepackage{amsmath}
\usepackage{hyperref}
\usepackage{balance}
\def\BibTeX{{\rm B\kern-.05em{\sc i\kern-.025em b}\kern-.08em
    T\kern-.1667em\lower.7ex\hbox{E}\kern-.125emX}}

\begin{document}

\title{DeepGate2: Functionality-Aware Circuit \\Representation Learning}

\author{
    \IEEEauthorblockN{Zhengyuan Shi\textsuperscript{1}, Hongyang Pan\textsuperscript{2}, Sadaf Khan\textsuperscript{1}, Min Li\textsuperscript{1}, Yi Liu\textsuperscript{1}, Junhua Huang\textsuperscript{3}, \\
    Hui-Ling Zhen\textsuperscript{3}, Mingxuan Yuan\textsuperscript{3}, Zhufei Chu\textsuperscript{2} and Qiang Xu\textsuperscript{1}}
    \IEEEauthorblockA{
        \textit{\textsuperscript{1} The Chinese University of Hong Kong, \textsuperscript{2} Ningbo University, \textsuperscript{3} Noah’s Ark Lab, Huawei} \\
        \{zyshi21, skhan, mli, yliu22, qxu\}@cse.cuhk.edu.hk \\
    }
    \vspace{-20pt}
}

\maketitle

\begin{abstract}
Circuit representation learning aims to obtain neural representations of circuit elements and has emerged as a promising research direction that can be applied to various EDA and logic reasoning tasks. Existing solutions, such as DeepGate, have the potential to embed both circuit structural information and functional behavior. However, their capabilities are limited due to weak supervision or flawed model design, resulting in unsatisfactory performance in downstream tasks.
In this paper, we introduce \textit{DeepGate2}, a novel functionality-aware learning framework that significantly improves upon the original DeepGate solution in terms of both learning effectiveness and efficiency. Our approach involves using pairwise truth table differences between sampled logic gates as training supervision, along with a well-designed and scalable loss function that explicitly considers circuit functionality. Additionally, we consider inherent circuit characteristics and design an efficient one-round graph neural network (GNN), resulting in an order of magnitude faster learning speed than the original DeepGate solution. Experimental results demonstrate significant improvements in two practical downstream tasks: logic synthesis and Boolean satisfiability solving.
The code is available at \href{https://github.com/cure-lab/DeepGate2}{https://github.com/cure-lab/DeepGate2}. 

\end{abstract}


\section{Introduction}
The application of Deep Learning (DL) techniques in Electronic Design Automation (EDA) has attracted a lot of attention, including routing~\cite{chen2020pros, mirhoseini2021graph}, synthesis~\cite{neto2019lsoracle, haaswijk2018deep}, and testing~\cite{shi2022deeptpi, huang2022neural}. Among learning-based EDA solutions, circuit representation learning~\cite{li2022deepgate, wang2022functionality, zhu2022tag, lai2022maskplace, fayyazi2019deep, he2021graph} has emerged as a promising research paradigm. This paradigm adopts a two-step approach instead of training individual models from scratch for each EDA task. First, a general model is pre-trained with task-agnostic supervision. Then, the model is fine-tuned for specific downstream tasks, resulting in improved performance.

One representative technique, called \emph{DeepGate}, embeds both the logic function and structural information of a circuit as vectors on each gate. DeepGate uses signal probabilities as the supervision task and applies an attention-based graph neural network (GNN) that mimics the logic computation procedure for learning. 
It has achieved remarkable results on testability analysis~\cite{shi2022deeptpi} and SAT solving problems~\cite{li2022deepsat}. However, we argue that its capabilities can still be dramatically enhanced. 


On one hand, DeepGate falls short in circuit functionality supervision. Generally speaking, the circuit truth table represents functionality in its most direct form. However, obtaining a complete truth table through exhaustive simulation is infeasible due to the exponentially increasing time requirement in relation to the number of primary inputs (PIs). DeepGate, as an alternative, employs the ratio of logic-1 in the truth table as a functionality-related supervision metric. It then approximates this value as the probability of logic-1 under randomized simulations performed a limited number of times. This approach, however, has a notable caveat. For example, a NOT gate and its fan-in gate can both have a logic-1 probability of $0.5$, but their truth tables are entirely opposite, thus highlighting the inadequacy of this supervision method.

On the other hand, DeepGate 
assigns the same initial embedding to all the PIs. Although these homogeneous embeddings reflect the equal logic probability of 0.5 for each PI under random simulation, they do not offer unique identifiers for individual PIs. Consequently, the model lacks the capacity to discern whether gates are reconvergent and driven by common PIs, information that is vital for circuit analysis. To preserve the logical correlation of gates, the model must execute multiple forward and backward propagation rounds to compensate for the absence of PI identification, which it achieves by revealing differences in local structure. Nevertheless, a model that involves many rounds of message-passing is time-consuming and inefficient, particularly when dealing with large circuits.
In response to these challenges, we present \emph{DeepGate2}, an innovative functionality-aware learning framework that notably advances the original DeepGate solution in both learning effectiveness and efficiency. Specifically, we incorporate the pairwise truth table difference of logic gates as supplementary supervision. This involves obtaining an incomplete truth table via rapid logic simulation, and then calculating the Hamming distance between the truth tables of two logic gates, referred to as the 'pairwise truth table difference'. Subsequently, we construct a functionality-aware loss function with the following objective: to minimize the disparity between the pairwise node embedding distance in the embedding space and the pairwise truth table difference, which serves as the ground truth. As a result, our proposed supervision introduces authentic functional information, a stark contrast to the initial DeepGate model, which predominantly depended on statistical facets of functionality.

Moreover, we introduce a single-round GNN architecture that efficiently encapsulates both structural and functional characteristics. Our GNN segregates the node embeddings into two discrete elements: functional embeddings and structural embeddings, each initialized differently. For the initial functional embeddings of the PIs, we assign a uniform vector to denote the equal logic probability shared among all PIs. For the initial structural embeddings of the PIs, we allocate a set of orthogonal vectors. These vectors, unique and uniformly spaced apart in the embedding space, mirror the functional independence of each PI. By transmitting these PI embeddings to the internal logic gates through two separate aggregation streams, our model effectively amalgamates both structural and functional data through a singular round of forward message-passing procedure.

We execute a range of experiments to highlight the effectiveness and efficiency of our proposed circuit representation learning framework. When compared to the predecessor, DeepGate, our model, DeepGate2, demonstrates a significant accuracy improvement and achieves an order of magnitude speedup in logic probability prediction. 
To further demonstrate the generalization capability of DeepGate2 in critical downstream applications that heavily rely on circuit functionality, we integrate it into EDA tools to aid logic synthesis~\cite{mishchenko2005fraigs} and SAT solving~\cite{queue2019cadical}. 
The experimental results further validate the efficacy of our DeepGate2 framework. 

The remainder of this paper is organized as follows. Section~\ref{Sec:Related} surveys related work. We then detail the proposed DeepGate2 framework in Section~\ref{Sec:Method}. We compare DeepGate2 with the original DeepGate and another functionality-aware solution~\cite{wang2022functionality} in Section~\ref{Sec:Experiment}. Next, we apply DeepGate2 onto several downstream tasks in Section~\ref{Sec:Task}. 
Finally, Section~\ref{Sec:Conclusion} concludes this paper. 

\section{Related Work}
\label{Sec:Related}
\subsection{Circuit Representation Learning}
A prominent trend in the deep learning community is to learn a general representation from data first and then apply it to various downstream tasks, for example, GPT~\cite{brown2020language} and BERT~\cite{devlin2018bert} learn representations of natural language text that can be fine-tuned for a wide range of natural language processing tasks. Circuit representation learning has also emerged as an attractive research direction, which falls into two categories: structure-aware circuit representation learning~\cite{zhu2022tag, he2021graph, fayyazi2019deep} and functionality-aware circuit representation learning~\cite{li2022deepgate, wang2022functionality}.

Since a circuit can be naturally formulated as a graph, with gates as nodes and wires as edge, the GNN is a powerful tool to capture the interconnections of logic gates and becomes a backbone model to learn circuit representations. For example, TAG~\cite{zhu2022tag} is a GNN-based model designed for analog and mixed-signal circuit representation learning and applied for several physical design applications, such as layout matching prediction, wirelength estimation, and net parasitic capacitance prediction. ABGNN~\cite{he2021graph} learns the representation of digital circuits and handles the arithmetic block identification task. However, these models tend to focus on structural encoding and are not suitable for functionality-related tasks.

Consequently, the functionality-aware circuit representation learning frameworks ~\cite{li2022deepgate, wang2022functionality} are designed to learn the underlying circuit functionality. 
For instance, FGNN~\cite{wang2022functionality} learns to distinguishes between functionally equivalent and inequivalent circuits by contrastive learning~\cite{wu2018unsupervised}. However, such self-supervised manner relies on data augmentation by perturbing the original circuit to logic equivalence circuit. If the perturbation is not strong and diverse, the model still identifies the functional equivalence circuits based on the invariant local structure, resulting a low generalization ability on capturing underlying functionality. 
DeepGate~\cite{li2022deepgate} leverages logic-1 probability under random simulation as supervision, which approximates the statistic of the most direct representation of functionality, i.e. truth table. Despite achieving remarkable progress on testability analysis~\cite{shi2022deeptpi}, there are limitations that affect the generalizability of DeepGate to other EDA tasks. We will elaborate on DeepGate in the next subsection.

\subsection{DeepGate Framework}
DeepGate~\cite{li2022deepgate} is the first circuit representation learning framework that embeds both structural and functional information of digital circuits. The model pre-processes the input circuits into a unified And-Inverter Graph (AIG) format and obtains rich gate-level representations, which can be applied to various downstream tasks. DeepGate treats the logic-1 probability as supervision to learn the functionality. Additionaly, the 
DeepGate consists of a GNN equipped with an attention-based aggregation function that propagates information of gates in levelized sequential manner.
The aggregation function learns to assign high attention weights to controlling fan-in of gates (e.g. the fan-in gate with logic-0 is the controlling fan-in of AND gate) that mimics the 
logic computation process. Although it has been applied to testability analysis~\cite{shi2022deeptpi} and SAT problem~\cite{li2022deepsat}, we argue that the model still encounters with two major shortcomings limiting its generalization ability. 


First, logic probability is not an appropriate supervision for learning functionality. The most direct representation of functionality is the truth table, however, using it as a training label is impractical due to the immeasurable computational overhead. DeepGate proposes to supervise the model by utilizing the proportion of logic-1 in the truth table and approximate this proportion as the logic probability through random simulation. 
However, logic probability is only a statistical information of functionality, indicating the number of logic-1 values in the truth table rather than which PI assignments lead to logic-1. Consequently, DeepGate cannot differentiate the functional difference between two circuits if they have the same probability.

\begin{figure}
    \centering
    \includegraphics[width=0.65\linewidth]{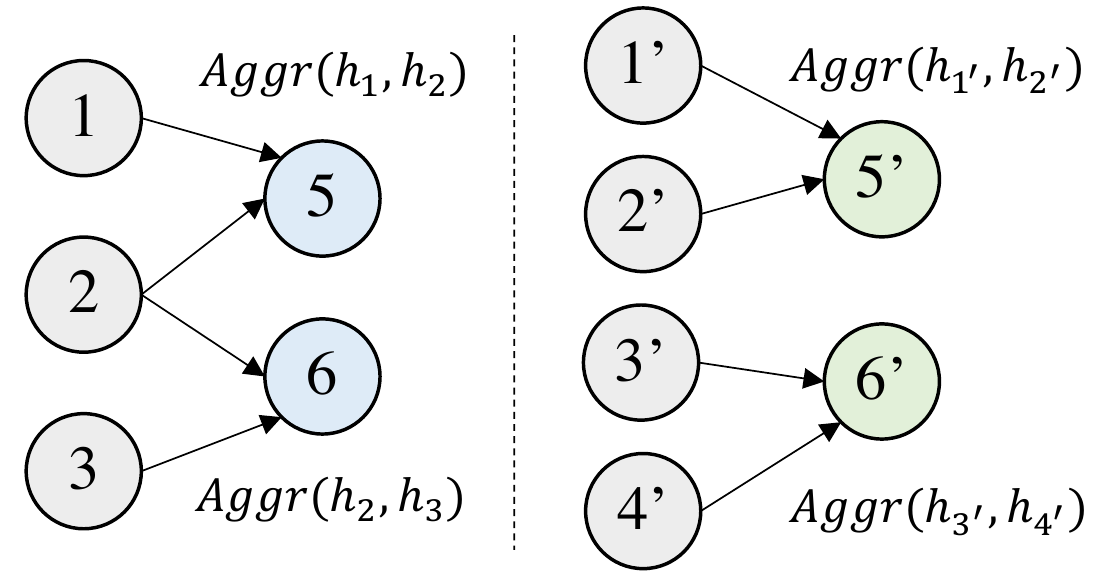}
    \vspace{-5pt}
    \caption{An example of reconvergence structure}
    \vspace{-15pt}
    \label{FIG:RC}
\end{figure}

Second, DeepGate is not efficient enough to deal with large circuit. Specifically, DeepGate requires to perform forward and backward message-passing operations for $20$ rounds to embed rich representations. Fig.~\ref{FIG:RC} illustrates the need of this multi-round GNN design in DeepGate where the nodes in grey color represent PIs. The incoming messages of nodes $5$, $6$, $5^{'}$, and $6^{'}$ during forward propagation are noted in the figure, where $h_i$ is the embedding vector of node $i$. Since, DeepGate uses the same initial embeddings for all nodes, the messages of nodes $5$, $6$, $5^{'}$, and $6^{'}$ in the first forward propagation round are identical. Thus, the model can only distinguish node embeddings based on their connections by repeatedly updating PIs through multiple rounds of forward and backward message propagation. 


We emphasize that the limitations of DeepGate comes from the lack of effective supervision and weak model design where the unique identification of all PIs are ignored. To address these issues, we propose an efficient one-round GNN design that maintains the unique identification of PIs and uses the pairwise truth-table difference of two gates as an effective supervision. 

\section{Methodology}
\label{Sec:Method}
\subsection{Problem Formulation}

The circuit representation learning model aims to map both circuit structure and functionality into embedding space, where the structure represents the connecting relationship of logic gates and the functionality means the logic computational mapping from inputs to outputs. We conclude that the previous models still lack of ability to capture functional information. In this paper, we propose to improve the previous DeepGate model~\cite{li2022deepgate} to represent circuits with similar functionality with the similar embedding vectors. In other words, these circuit representations should have short distance in the embedding space. 

\begin{figure} [!t]
    \centering
    \includegraphics[width=1.0\linewidth]{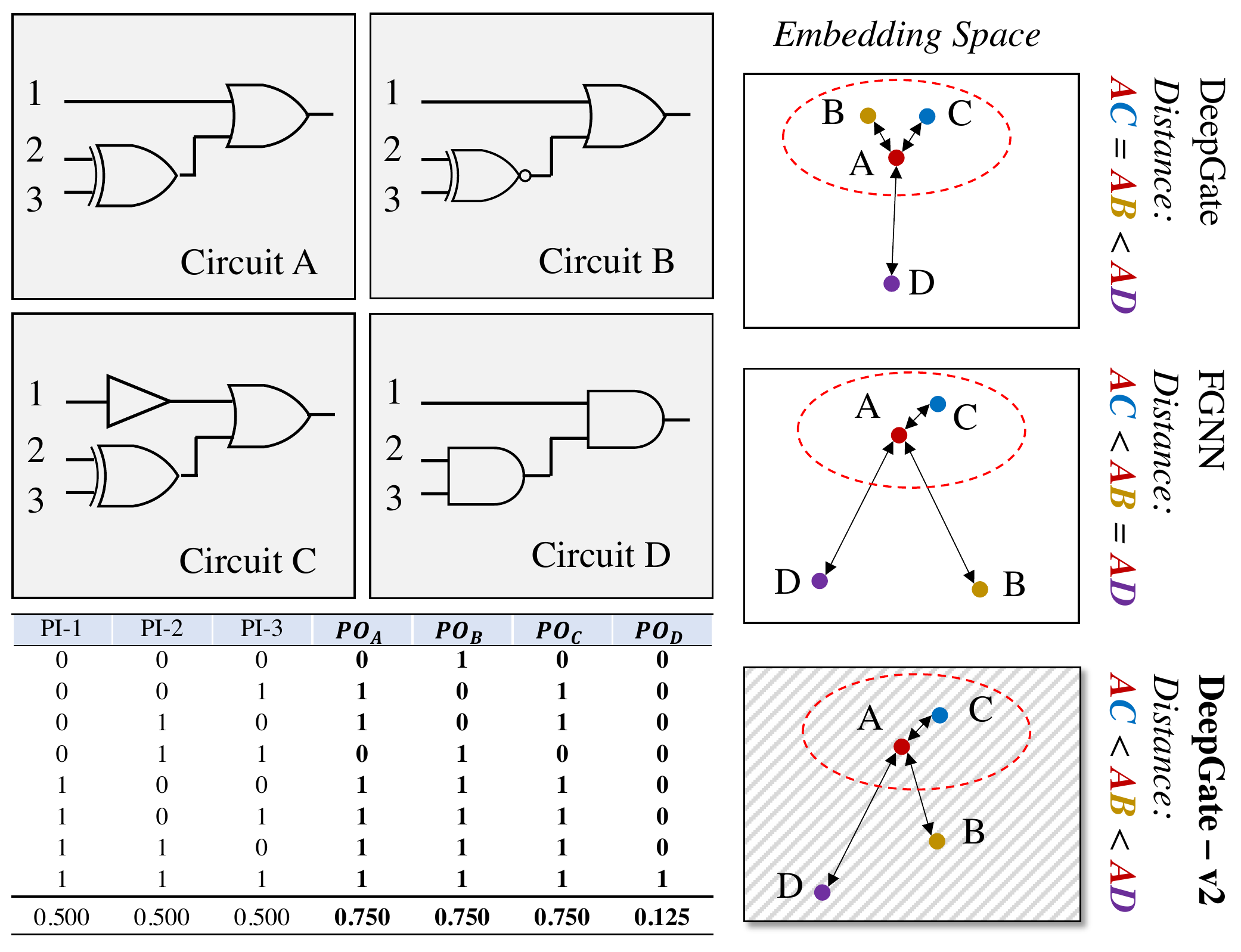}
    \caption{Problem statement: the embedding vectors should be close if circuit functions are similar}
    \label{fig:problem}
    \vspace{-8pt}
\end{figure}

We take Circuit A, B, C, and D as examples in Fig.~\ref{fig:problem}, where all of them have similar topological structures. Since Circuit A, B and C perform with the same logic probability, DeepGate~\cite{li2022deepgate} tends to produce the similar embeddings for these three circuits. Hence, it is hard to identify the logic equivalent circuits by DeepGate. Although FGNN~\cite{wang2022functionality} is trained to classify logical equivalence and inequivalence circuits by contrastive learning, they cannot differentiate the relative similarity. As shown in the embedding space, the distance between A and B is equal to the distance between A and D. Nonetheless, as indicated in the truth table, Circuit A is equivalent to Circuit C, similar to Circuit B (with only $2$ different bits), but dissimilar to Circuit D (with $5$ different bits).

We expect that the model will bring together or separate the circuits in embedding space according to their truth tables. Therefore, the expected DeepGate2 model not only identifies the logic equivalent nodes, but also predicts the functional similarity. Thus, we can apply such functionality-aware circuit learning model to provide benefits for the real-world applications.


\subsection{Dataset Preparation} \label{Sec:Method:Data}
To train the circuit representation learning model, we need to find a supervision contains rich functional information and prepare an effective dataset at first. \textit{Truth table}, which records the complete logic computational mapping,  provides the most direct supervision. However, the length of the truth table increases exponentially with the number of primary inputs, and obtaining a complete truth table requires an immeasurable amount of time. Therefore, a reasonable supervision should be easily obtained and closely related to the truth table.

Firstly, we use the Hamming distance between truth tables of two logic gates as supervision. That is, in a way similar to metric learning~\cite{hoffer2015deep}, we map nodes to an embedding space and hope that the distance of the embedding vectors is positive correlated with the Hamming distance of the truth table. Formally, we denote the truth table vector of node $i$ is $T_i$ and the embedding vector of node $i$ is $h_i$. 
\begin{equation} \label{Eq:propto}
    distance(h_i, h_j) \propto distance(T_i, T_j)
\end{equation} 

Secondly, to improve the data efficiency, we regard the each logic gate in circuit as a new circuit (logic cone) with the current gate as output and the original PIs as inputs. By parsing a single original circuit, we obtain a large number of new circuits. Therefore, the task of graph learning becomes the task of learning node-level representation, and the difficulty of data collection is reduced.

Thirdly, to ensure the quality of sample pairs and limit the number of sample pairs, we impose the following constraints during sampling node pairs: (1) Two logic cones of the two nodes should have the same PI, which is a necessary condition for comparing the truth table difference. (2) The logic probability, which is the number of logic-1 percentage in the truth table, should be similar (distance within $5\%$). This is because if the logic probability of two nodes is not consistent, their functions are definitely not consistent. If the logic probability of two nodes is consistent, their functions may be consistent. (3) The difference in logic levels between two nodes should be within $5$, because when the two nodes are far apart, their functions are unlikely to be correlated. (4) We only consider the extreme cases, namely, the difference between truth tables is within $20\%$ or above $80\%$.

We do not perform the complete simulation, but set a maximum simulation time to obtain the response of each node as an incomplete truth table. It should be noted that we utilize the And-Inverter Graph (AIG) as the circuit netlist format, which is only composed of AND gate and NOT gate. Any other logic gates, including OR, XOR and MUX, can be transformed into a combination of AND and NOT gates in linear time. 

\subsection{Functionality-Aware Loss Function} \label{Sec:Method:Func}
The primary objective of our purposed functionality-aware circuit learning model is to learn node embeddings, where two embedding vectors will be similar if the corresponding two node function are similar. As we sample node pairs $\mathcal{N}$ in the Section~\ref{Sec:Method:Data}, we can obtain the Hamming distance of truth table $D^T$ of each node pair. 
\begin{equation}
    D_{(i, j)}^T = \frac{HammingDistance(T_i, T_j)}{length(T_i)}, (i, j) \in \mathcal{N}
\end{equation}

According to Eq.~\eqref{Eq:propto}, the distance of embedding vectors $D^H$ should be proportional to the Hamming distance of the truth table $D^T$. We define the distance of embedding vectors in Eq.~\eqref{Eq:Cos}, where is calculated based on cosine similarity. In other word, the similarity of embedding vectors $S_{(i, j)}$ should be negative related to distance $D_{(i, j)}^T$.
\begin{equation} \label{Eq:Cos}
    \begin{split}
        S_{(i, j)} & = CosineSimilarity(h_i, h_j) \\ 
        D_{(i, j)}^H & = 1 - S_{(i, j)}
    \end{split}
\end{equation}

Therefore, the training objective is to minimize the difference between $D^H$ and $D^T$. We purpose the functionality-aware loss function $L_{func}$ as below. 
\begin{equation}
    \begin{split}
        D_{(i, j)}^{T'} & = ZeroNorm(D_{(i, j)}^T) \\ 
        D_{(i, j)}^{H'} & = ZeroNorm(D_{(i, j)}^H) \\ 
        L_{func} & = \sum_{(i, j) \in \mathcal{N}} (L1Loss(D_{(i, j)}^{T'}, D_{(i, j)}^{H'}))
    \end{split}
\end{equation}

\subsection{One-round GNN Model} \label{Sec:Method:Model}
In this subsection, we propose a GNN model that can capture both functional and structural information for each logic gate through one-round forward propagation. 

First, we propose to separate the functional embeddings $hf$ and structural embeddings $hs$, and initialize them in difference ways. We assign the uniform initial functional embeddings to primary inputs (PI), as they all have equivalent logic probability under random simulation. However, we design a PI encoding (PIE) strategy by assigning a unique identification to each PI as its initial structural embedding. Specifically, the initial PI structural embeddings $hs_{i}, i \in PI$ are orthogonal vectors. This means that the dot product of any two PIs' embeddings is zero.

Second, we design four aggregators: $aggr_{AND}^{s}$ aggregates the message for structural embedding $hs$ of an AND gate, $aggr_{AND}^{f}$ aggregates the message for functional embedding $hf$ of an AND gate. $aggr_{NOT}^{s}$ and $aggr_{NOT}^{f}$ update $hs$ and $hf$ of a NOT gate, respectively. 

We implement each aggregator using the self-attention mechanism~\cite{vaswani2017attention}, as the output of a logic gate is determined by the controlling values of its fan-in gates. For example, an AND gate must output logic-0 if any of its fan-in gates has logic-0. By employing the attention mechanism, the model learns to assign greater importance to the controlling inputs~\cite{li2022deepgate}. As illustrated in Eq.~\eqref{Eq:aggr}, $w_q$, $w_k$ and $w_v$ are three weight matrices and $d$ is the dimension of embedding vectors $h$. 
\begin{equation} \label{Eq:aggr}
    \begin{split} 
        \alpha_j & = softmax(\frac{w_q^\top h_i \cdot (w_k^\top h_j)^\top}{\sqrt{d}}) \\ 
        m_j & = w_v^\top h_j \\ 
        h_i & = aggr(h_j | j \in \mathcal{P}(i)) = \sum_{j \in \mathcal{P}(i)}(\alpha_j * m_j)  
    \end{split}
\end{equation}

Third, during forward propagation, the structural embeddings are updated only with the structural embeddings of predecessors. As shown in Eq.~\eqref{Eq:hs}, where the Gate $a$ is AND gate, the Gate $b$ is NOT gate.  
\begin{equation} \label{Eq:hs}
    \begin{split}
        hs_{a} & = aggr_{AND}^{s} (hs_j | j \in \mathcal{P}(a)) \\
        hs_{b} & = aggr_{NOT}^{s} (hs_j | j \in \mathcal{P}(b)) 
    \end{split}
\end{equation}

At the same time, the gate function is determined by the function and the structural correlations of the fan-in gates. Therefore, the functional embeddings are updated as Eq.~\eqref{Eq:hf}. 
\begin{equation} \label{Eq:hf}
    \begin{split}
        hf_{a} & = aggr_{AND}^{f} ([hs_j, hf_j] | j \in \mathcal{P}(a)) \\ 
        hf_{b} & = aggr_{NOT}^{f} ([hs_j, hf_j] | j \in \mathcal{P}(b)) \\ 
    \end{split}
\end{equation}

\begin{figure} [!t]
    \centering
    \includegraphics[width=0.6\linewidth]{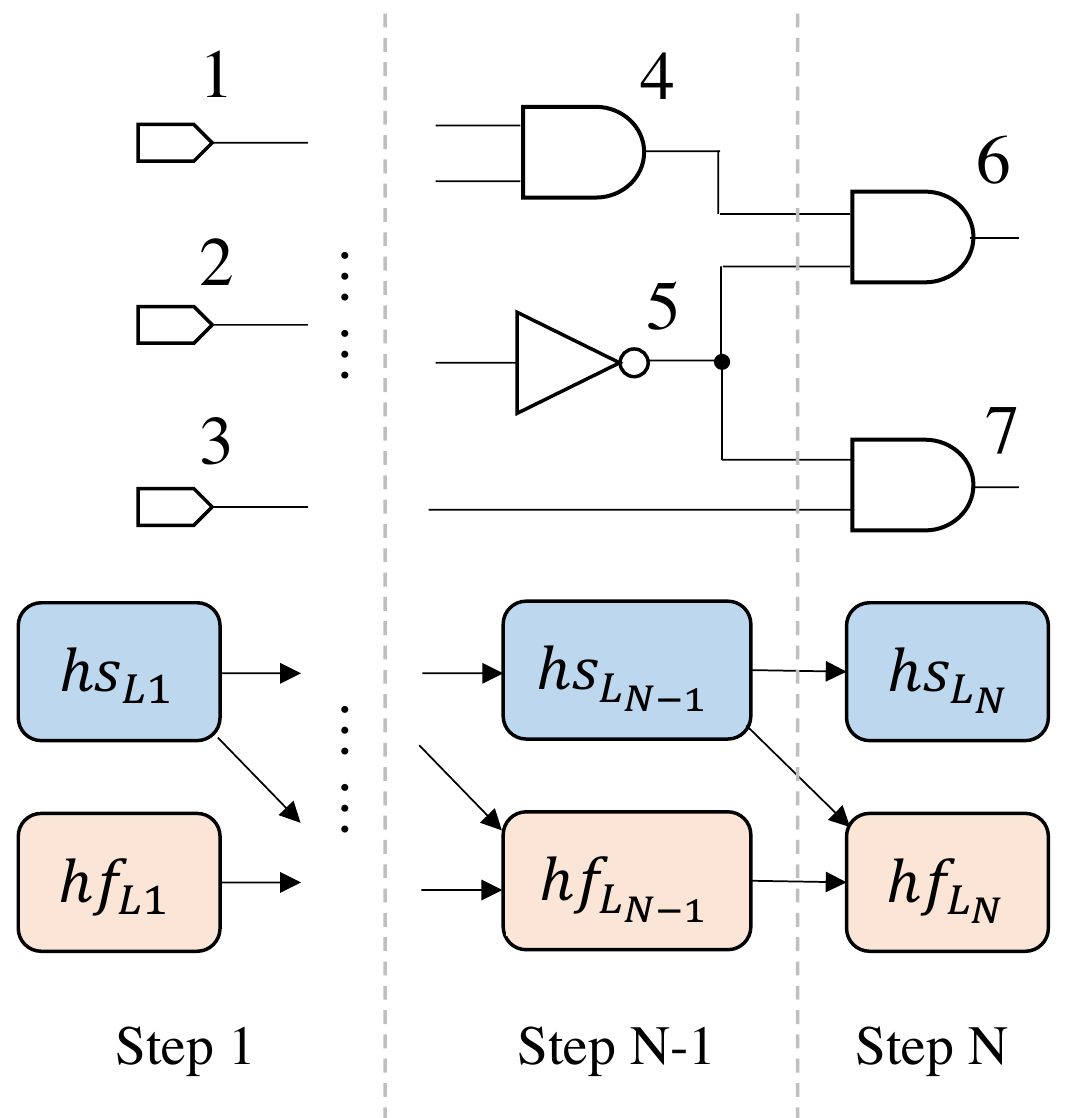}
    \vspace{-5pt}
    \caption{One-round GNN propagation process}
    \vspace{-10pt}
    \label{FIG:GNNProc}
\end{figure}

Therefore, as shown in Fig.~\ref{FIG:GNNProc}, the GNN propagation process performs from PI to PO level by level. For the node in level $l$, its structural embedding $hs_{L_{l}}$ will be updated with the structural embeddings of the node in level $l-1$. Additionally, the functional embedding $hf_{L_{l}}$ will be updated with both structural embeddings $hs_{L_{l-1}}$ and functional embeddings $hf_{L_{l-1}}$. The GNN propagation completes after processing $N$ levels. 

\begin{figure*} [!t]
    \centering
    \includegraphics[width=0.98\linewidth]{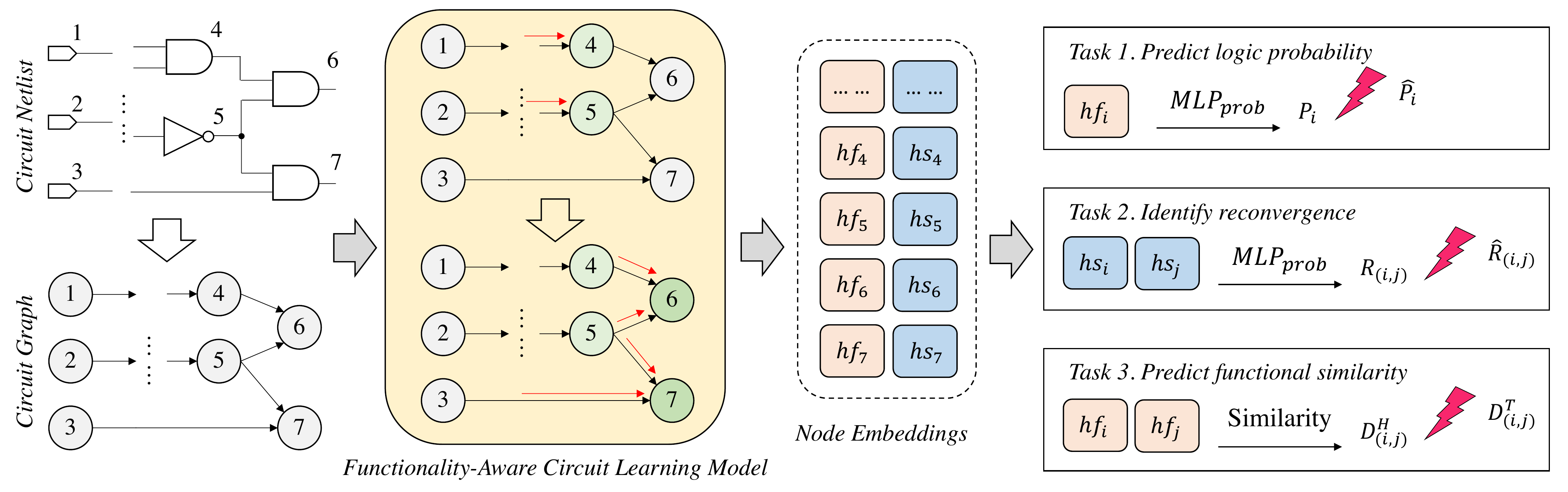}
    \caption{Functionality-aware circuit learning framework}
    \label{fig:model}
\end{figure*}

\subsection{Model Training Strategies}
To train the model, we employed multi-stage training strategy, similar to training a model with an easy task and then a harder task in curriculum learning~\cite{bengio2009curriculum}. During each stage, we trained the model with multiple supervisions in multi-task learning manner~\cite{ruder2017overview}. 
In the first stage, we train the one-round GNN model with two simple tasks. The Task 1 involves predicting the logic probability, while the Task 2 entails identifying the structural correlation. To achieve this, we readout the functional embedding $hf_i$ to predict the logic probability $\hat{P_i}$ by a multi-layer perceptron (MLP), denoted as $MLP_{prob}$. In addition, we utilize the structural embeddings $hs_i$ and $hs_j$ to predict whether node $i$ and node $j$ can be reconvergent by $MLP_{rc}$. 
\begin{equation}
    \begin{split}
        \hat{P_i} & = MLP_{prob}(hf_i) \\
        \hat{R_{(i, j)}} & = MLP_{rc}(hs_i, hs_j) 
    \end{split}
\end{equation}

We define the loss function for Task 1 in Eq.~\eqref{Eq:LossProb}, where the $P_i$ is the ground truth logic probability obtained through random simulation. 
\begin{equation} \label{Eq:LossProb}
    L_{prob} = L1Loss(P_i, \hat{P_i})
\end{equation}

Besides, we define the loss function for Task 2 in Eq.~\eqref{Eq:LossRC}. The binary ground truth, denoted as $R_{(i, j)}$, indicates whether node pair $i$ and $j$ have a common predecessor. 
\begin{equation} \label{Eq:LossRC}
    L_{rc} = BCELoss(R_{(i, j)}, \hat{R_{(i, j)}})
\end{equation}

Consequently, the loss function for Stage 1 is presented in Eq.~\eqref{Eq:Stage1Loss}, where the $w_{prob}$ and $w_{rc}$ are the weight for Task 1 and Task 2, respectively.  
\begin{equation} \label{Eq:Stage1Loss}
    L_{stage1} = L_{prob} * w_{prob} + L_{rc} * w_{rc}
\end{equation}

The second training stage involves another more difficult Task 3. functionality-aware learning, as described in Section~\ref{Sec:Method:Func}. The loss function for Stage 2 is defined below, where $w_{func}$ represents the loss weight of Task 3.
\begin{equation}
    L_{stage2} = L_{prob} \times w_{prob} + L_{rc} \times w_{rc} + L_{func} \times w_{func}
\end{equation}

Overall, the model can differentiate gates with varying probability in Stage 1. As the logic equivalent pairs only occur when nodes have the same probability, the model in Stage 2 learns to predicting the functional similarity within the probability equivalent class. The effectiveness of the above training strategies is demonstrated in Section~\ref{Sec:Exp:Train}. 

\vspace{-10pt}

\section{Experiments}
\label{Sec:Experiment}
In this section, we demonstrate the ability of our proposed DeepGate2 to learn functionality-aware circuit representations. Firstly, Section~\ref{Sec:Exp:Setting} provides the preliminary of our experiments, including details on dataset preparation, evaluation metrics and model settings. Secondly, we compare the effectiveness and efficiency of our DeepGate2 against DeepGate~\cite{li2022deepgate} and FGNN~\cite{wang2022functionality} on two function-related tasks: logic probability prediction (see Section~\ref{Sec:Exp:Prob}) and logic equivalence gates identification (see Section~\ref{Sec:Exp:EqGate}). Thirdly, we investigate the effectiveness of model design and training strategies in Section~\ref{Sec:Exp:Model} and Section~\ref{Sec:Exp:Train}, respectively. 

\subsection{Experiment Settings} \label{Sec:Exp:Setting}
\subsubsection{Dataset Preparation}
We use the circuits in DeepGate~\cite{li2022deepgate}, which are extracted from ITC’99~\cite{ITC99}, IWLS’05~\cite{albrecht2005iwls}, EPFL~\cite{EPFLBenchmarks} and OpenCore~\cite{takeda2008opencore}. These circuits consists of $10,824$ AIGs with sizes ranging from $36$ to $3,214$ logic gates. 
To obtain the incomplete truth table, we generate $15,000$ random patterns and record the corresponding response. Following the data preparation method described in Section~\ref{Sec:Method:Data}, we construct a dataset comprising $894,151$ node pairs. We create 80/20 training/test splits for model training and evaluation. 

\subsubsection{Evaluation Metrics} \label{Sec:Exp:Acc}
We assess our DeepGate2 with two tasks. The first task is to predict the logic probability for each logic gate. We calculate the average prediction error (PE) as Eq.~\eqref{Eq:AccTask0Prob}, where the set $\mathcal{V}$ includes all logic gates. 
\begin{equation} \label{Eq:AccTask0Prob}
    PE = \frac{1}{|\mathcal{V}|} \sum_{i \in \mathcal{V}}|P_i - \hat{P}_i|
\end{equation}

The second task is to identify the logic equivalence gates within a circuit. A gate pair $(i, j)$ is considered as a positive pair if these two logic gates $i$ and $j$ have the same function, where the pairwise Hamming distance of truth tables $D_{(i, j)}^T = 0$. If the similarity $S_{(i, j)}$ between these two embedding vectors $hf_i$ and $hf_j$ exceeds a certain threshold, the model will recognize the gate pair $(i, j)$ as equivalent. The optimal threshold $\theta$ is determined based on the receiver operating characteristic (ROC). The evaluation metrics is formally defined in Eq.~\eqref{Eq:AccTask1}, where $TP$, $TN$, $FP$, $FN$ are true positive, true negative, false positive and false negative, respective, and $M$ is the total number of gate pairs. In the following experiments, the performance on logic equivalence gates identification is measured in terms of \text{Recall}, \text{Precision}, \text{F1-Score} and area under curve (AUC). 
\begin{equation} \label{Eq:AccTask1}
    \begin{split}
        TP & = \frac{\sum( (D_{(i, j)}^T == 0) \ \& \ (S_{(i, j)} > \theta))}{M} \\
        TN & = \frac{\sum( (D_{(i, j)}^T > 0) \ \& \ (S_{(i, j)} < \theta))}{M} \\
        FP & = \frac{\sum( (D_{(i, j)}^T > 0) \ \& \ (S_{(i, j)} > \theta))}{M} \\
        FN & = \frac{\sum( (D_{(i, j)}^T == 0) \ \& \ (S_{(i, j)} < \theta))}{M} \\
    \end{split}
\end{equation}

We conduct the following performance comparisons on $10$ industrial circuits, with circuit sizes ranging from $3.18k$ gates to $40.50k$ gates. 

\subsubsection{Model Settings}
In the one-round GNN model configuration, the dimension of both structural embedding $hs$ and functional embedding $hf$ is $64$. Both $MLP_{prob}$ and $MLP_{rc}$ contain $1$ hidden layer with $32$ neurons and a ReLu activation function. 

The model is trained for $60$ epochs to ensure each model can converge. The other models~\cite{li2022deepgate, wang2022functionality} mentioned in the following experiments maintain their original settings and are trained until they converge. 
We train all the models for $80$ epochs with batch-size $16$ on a single Nvidia V100 GPU. We adopt the Adam optimizer~\cite{kingma2014adam} with learning rate $10^{-4}$ and weight decay $10^{-10}$.

\subsection{Comparison with DeepGate on Probability Prediction} \label{Sec:Exp:Prob}
We compare the probability prediction error (PE, see Eq.~\eqref{Eq:AccTask0Prob}) and runtime (Time) with previous DeepGate. The previous DeepGate is denoted as \textit{DeepGate} and our proposed model with novel loss function and GNN design is named as \textit{DeepGate2} in Table~\ref{TAB:Comp:TaskProb}. Based on the results presented in the table, we make two observations.
First, our proposed DeepGate2 exhibits more accurate predictions of logic probability compared to the previous version. On average, the probability prediction error (PE) of DeepGate2 is $13.08\%$ lower than that of DeepGate. This suggests that using the novel model architecture with embedding initialization strategy can benefit logic representation learning and lead to better results on logic probability prediction. 
Second, our DeepGate2 performs more efficient than DeepGate. Take the circuit D1 as an example, DeepGate requires $36.89$ seconds for inference, but our DeepGate2 only needs $2.23$s, which is $16.56$x faster than the previous DeepGate. Moreover, compared to the previous model, DeepGate2 achieves an order of magnitude speedup ($16.43$x on average) in model runtime. This is attributed to the fact that the GNN model in DeepGate relies on $10$ forward and $10$ backward message propagation, whereas the proposed one-round GNN model in DeepGate2 only performs forward propagation for $1$ time. 
Therefore, the new circuit representation learning model is more effective and efficient than DeepGate, and demonstrates the generalization ability on large-scale circuits. 

\begin{table}[!t]
\centering
\caption{Performance of DeepGate (v1) and our proposed DeepGate2 (v2) on Logic Probability Prediction} \label{TAB:Comp:TaskProb}
\renewcommand\tabcolsep{4.0pt}
\begin{tabular}{ll|ll|ll|ll}
\toprule
Circuit       & \#Gates   & \multicolumn{2}{c|}{DeepGate} & \multicolumn{2}{c|}{DeepGate2} & \multicolumn{2}{c}{Reduction}     \\
              &           & PE            & Time      & PE             & Time          & PE               & Time(x)       \\ \midrule
D1            & 19,485    & 0.0344        & 36.89s         & 0.0300         & 2.23s          & 12.79\%          & 16.56          \\
D2            & 12,648    & 0.0356        & 16.11s        & 0.0309         & 1.18s          & 13.20\%          & 13.66          \\
D3            & 14,686    & 0.0355        & 21.42s        & 0.0294         & 1.44s          & 17.18\%          & 14.92          \\
D4            & 7,104     & 0.0368        & 5.89s         & 0.0323         & 0.50s          & 12.23\%          & 11.89          \\
D5            & 37,279    & 0.0356        & 131.22s        & 0.0316         & 7.82s          & 11.24\%          & 16.79          \\
D6            & 37,383    & 0.0325        & 133.02s        & 0.0285         & 8.10s          & 12.31\%          & 16.42          \\
D7            & 10,957    & 0.0357        & 13.02s         & 0.0316         & 0.90s          & 11.48\%          & 14.44          \\
D8            & 3,183     & 0.0406        & 1.60s          & 0.0341         & 0.17s          & 16.01\%          & 9.64           \\
D9            & 27,820    & 0.0368        & 77.20s         & 0.0322         & 4.73s          & 12.50\%          & 16.34          \\
D10           & 40,496    & 0.0327        & 154.00s        & 0.0290         & 8.89s          & 11.31\%          & 17.33          \\ \midrule
\textbf{Avg.} & \textbf{} & \textbf{0.0356} & \textbf{59.04s} & \textbf{0.0310} & \textbf{3.59s} & \textbf{13.08\%} & \textbf{16.43} \\ \bottomrule
\end{tabular}
\vspace{-10pt}
\end{table}

\subsection{Comparison with other Models on Logic Equivalence Gates Identification} \label{Sec:Exp:EqGate}
This section compares the functionality-aware accuracy, as defined in Section~\ref{Sec:Exp:Acc} of DeepGate2 with that of two other models: DeepGate~\cite{li2022deepgate} and FGNN~\cite{wang2022functionality}. The DeepGate~\cite{li2022deepgate} model treats the logic probability as supervision since it contains the statistical information of truth table. The FGNN~\cite{wang2022functionality} is trained to differentiate between logic equivalent and inequivalent circuits using contrastive learning. 

\begin{table*}[!t]
\centering
\caption{Performance of Different Models on Logic Equivalence Gates Identification} \label{TAB:Comp:Task1}
\vspace{-5pt}
\begin{tabular}{@{}ll|lll|lll|lll@{}}
\toprule
\multicolumn{1}{c}{\multirow{2}{*}{Circuit}} & \multicolumn{1}{c|}{\multirow{2}{*}{Size}} & \multicolumn{3}{c|}{FuncModel}                                                             & \multicolumn{3}{c|}{DeepGate}                                                              & \multicolumn{3}{c}{FGNN}                                                                  \\
\multicolumn{1}{c}{}                         & \multicolumn{1}{c|}{}                      & \multicolumn{1}{c}{Recall} & \multicolumn{1}{c}{Precision} & \multicolumn{1}{c|}{F1-Score} & \multicolumn{1}{c}{Recall} & \multicolumn{1}{c}{Precision} & \multicolumn{1}{c|}{F1-Score} & \multicolumn{1}{c}{Recall} & \multicolumn{1}{c}{Precision} & \multicolumn{1}{c}{F1-Score} \\ \midrule
D1                                           & 19,485                                     & 98.46\%                    & 84.34\%                       & 0.9085                        & 94.59\%                    & 52.69\%                       & 0.6768                        & 63.64\%                    & 41.18\%                       & 0.5000                       \\
D2                                           & 12,648                                     & 99.39\%                    & 89.07\%                       & 0.9395                        & 92.07\%                    & 52.98\%                       & 0.6726                        & 60.87\%                    & 35.00\%                       & 0.4444                       \\
D3                                           & 14,686                                     & 97.89\%                    & 84.36\%                       & 0.9062                        & 93.25\%                    & 62.08\%                       & 0.7454                        & 72.22\%                    & 44.83\%                       & 0.5532                       \\
D4                                           & 7,104                                      & 98.90\%                    & 97.83\%                       & 0.9836                        & 90.11\%                    & 46.33\%                       & 0.6120                        & 62.73\%                    & 23.53\%                       & 0.3422                       \\
D5                                           & 37,279                                     & 99.80\%                    & 88.49\%                       & 0.9381                        & 93.69\%                    & 56.72\%                       & 0.7066                        & 64.00\%                    & 69.57\%                       & 0.6667                       \\
D6                                           & 37,383                                     & 98.31\%                    & 93.19\%                       & 0.9568                        & 93.23\%                    & 60.49\%                       & 0.7337                        & 59.09\%                    & 46.67\%                       & 0.5215                       \\
D7                                           & 10,957                                     & 99.15\%                    & 97.48\%                       & 0.9831                        & 88.89\%                    & 48.83\%                       & 0.6303                        & 36.36\%                    & 23.53\%                       & 0.2857                       \\
D8                                           & 3,183                                      & 97.14\%                    & 97.14\%                       & 0.9714                        & 82.86\%                    & 49.90\%                       & 0.6229                        & 62.63\%                    & 22.22\%                       & 0.3280                       \\
D9                                           & 27,820                                     & 99.74\%                    & 85.16\%                       & 0.9188                        & 92.51\%                    & 48.77\%                       & 0.6387                        & 62.50\%                    & 26.79\%                       & 0.3750                       \\
D10                                          & 40,496                                     & 98.48\%                    & 87.80\%                       & 0.9283                        & 93.35\%                    & 61.22\%                       & 0.7395                        & 47.02\%                    & 32.63\%                       & 0.3853                       \\ \midrule
\textbf{Avg.}                                & \textbf{}                                  & \textbf{98.73\%}           & \textbf{90.49\%}              & \textbf{0.9434}               & \textbf{91.46\%}           & \textbf{54.00\%}              & \textbf{0.6778}               & \textbf{59.11\%}           & \textbf{36.60\%}              & \textbf{0.4402}              \\ \bottomrule
\end{tabular}
\vspace{-5pt}
\end{table*}

Table~\ref{TAB:Comp:Task1} presents the performance of three models on the task of logic equivalence gates identification. 
Firstly, our proposed approach outperforms the other two models on all circuits with an average F1-score of $0.9434$, while DeepGate and FGNN only achieve F1-Score $0.6778$ and $0.4402$, respectively. For instance, in circuit D7, our proposed functionality-aware circuit learning approach achieves an F1-Score of $0.9831$ and accurately identifies $99.15\%$ of logic equivalence gate pairs with a precision of $97.48\%$, indicating a low false positive rate. In contrast, DeepGate only achieves an F1-score of $0.6778$, while FGNN fails on most of the pairs. 
Secondly, although DeepGate has an average recall of $91.46\%$, its precision is only $54.00\%$, indicating a large number of false positive identifications. This is because DeepGate can only identify logic equivalent pairs by predicting logic probability, which leads to incorrect identification of gate pairs with similar logic probability. According to our further experiment, in $80.83\%$ of false positive pairs, the model incorrectly identifies gate pairs with similar logic probability as functionally equivalent. 
Thirdly, FGNN achieves the lowest performance among the other models, with only $0.4402$ F1-Score. The poor performance of FGNN is attributed to the lack of effective supervision. While FGCN learns to identify logic equivalence circuits generated by perturbing local structures slightly, the model tends to consider circuits with similar structures to have the same functionality. However, in the validation dataset and practical applications, two circuits may have the same function even if their topological structures are extremely different. Therefore, the self-supervised approach limits the effectiveness of FGNN in identifying logic equivalence gates.

\subsection{Effectiveness of PI Encoding Strategy} \label{Sec:Exp:Model}
To demonstrate the effectiveness of our proposed PI encoding (PIE) strategy, we trained another model without assigning unique identifications for PIs, which we refer to as \textit{w/o PIE}. The results are presented in Table~\ref{TAB:ABS:PIE}, which show that disabling the PIE reduces the F1-Score of identifying logic equivalence gates from $0.9434$ to $0.7541$, resulting in an average reduction of $20.07\%$.

\begin{table}[!t]
\centering
\caption{Performance Comparison between w/ PIE and w/o PIE on Logic Equivalence Gates Identification} 
\vspace{-5pt} \label{TAB:ABS:PIE}
\renewcommand\tabcolsep{4.0pt}
\begin{tabular}{@{}l|lll|lll@{}}
\toprule
\multicolumn{1}{c|}{\multirow{2}{*}{Circuit}} & \multicolumn{3}{c|}{w/ PIE}                                                                & \multicolumn{3}{c}{w/o PIE}                                                               \\
\multicolumn{1}{c|}{}                         & \multicolumn{1}{c}{Recall} & \multicolumn{1}{c}{Precision} & \multicolumn{1}{c|}{F1-Score} & \multicolumn{1}{c}{Recall} & \multicolumn{1}{c}{Precision} & \multicolumn{1}{c}{F1-Score} \\ \midrule
D1                                            & 98.46\%                    & 84.34\%                       & 0.9085                        & 70.66\%                    & 64.66\%                       & 0.6753                       \\
D2                                            & 99.39\%                    & 89.07\%                       & 0.9395                        & 87.20\%                    & 78.57\%                       & 0.8266                       \\
D3                                            & 97.89\%                    & 84.36\%                       & 0.9062                        & 79.15\%                    & 76.21\%                       & 0.7765                       \\
D4                                            & 98.90\%                    & 97.83\%                       & 0.9836                        & 87.91\%                    & 59.70\%                       & 0.7111                       \\
D5                                            & 99.80\%                    & 88.49\%                       & 0.9381                        & 79.23\%                    & 76.73\%                       & 0.7796                       \\
D6                                            & 98.31\%                    & 93.19\%                       & 0.9568                        & 87.10\%                    & 77.56\%                       & 0.8205                       \\
D7                                            & 99.15\%                    & 97.48\%                       & 0.9831                        & 84.62\%                    & 64.13\%                       & 0.7296                       \\
D8                                            & 97.14\%                    & 97.14\%                       & 0.9714                        & 74.29\%                    & 83.87\%                       & 0.7879                       \\
D9                                            & 99.74\%                    & 85.16\%                       & 0.9188                        & 62.27\%                    & 76.27\%                       & 0.6856                       \\
D10                                           & 98.48\%                    & 87.80\%                       & 0.9283                        & 87.07\%                    & 65.54\%                       & 0.7479                       \\ \midrule
\textbf{Avg.}                                 & \textbf{}                  & \textbf{}                     & \textbf{0.9434}               & \textbf{}                  & \textbf{}                     & \textbf{0.7541}              \\ \bottomrule
\end{tabular}
\end{table}

Such reduction can be attributed to the fact that, as demonstrated as the failure case in Section~\ref{Sec:Related} and Fig.~\ref{FIG:RC}, the one-round GNN model without the PIE strategy cannot model the structural information of the circuit. More specifically, the accuracy of the reconvergence structure identification task with w/ PIE model is $93.22\%$, while the w/o model only achieve $74.56\%$. The functionality of logic gate is affected by both functionality of fan-in gates and whether there is reconvergence between its fan-in gates. Once the reconvergence structure cannot be accurately identified, node functionality cannot be modeled accurately.

\subsection{Effectiveness of Training Strategies} \label{Sec:Exp:Train}
To investigate the effectiveness of our multi-stage training strategy, we train another model (noted as w/o multi-stage model) with all loss functions in only one stage, instead of adding the functionality-aware loss function in the second stage. The original model with multiple stages training strategy is noted as w/ multi-stage model. 
The w/ multi-stage model learn to predict the logic probability and structural correlation in the first stage and learn the more difficult task, which predicts the functionality in the second stage. The results are shown in Table~\ref{TAB:ABS:multi-stage}, where the model w/ multi-stage achieves an F1-Score of $0.9434$ on average and the model w/o multi-stage achieves only $0.7137$. 

\begin{table}[!t]
\centering
\caption{Performance Comparison between w/ multi-stage and w/o multi-stage on Logic Equivalence Gates Identification} \label{TAB:ABS:multi-stage}
\renewcommand\tabcolsep{4.0pt}
\begin{tabular}{@{}l|lll|lll@{}}
\toprule
\multicolumn{1}{c|}{\multirow{2}{*}{Circuit}} & \multicolumn{3}{c|}{w/ multi-stage}                                                        & \multicolumn{3}{c}{w/o multi-stage}                                                       \\
\multicolumn{1}{c|}{}                         & \multicolumn{1}{c}{Recall} & \multicolumn{1}{c}{Precision} & \multicolumn{1}{c|}{F1-Score} & \multicolumn{1}{c}{Recall} & \multicolumn{1}{c}{Precision} & \multicolumn{1}{c}{F1-Score} \\ \midrule
D1                                            & 98.46\%                    & 84.34\%                       & 0.9085                        & 79.46\%                    & 67.68\%                       & 0.7310                       \\
D2                                            & 99.39\%                    & 89.07\%                       & 0.9395                        & 74.39\%                    & 63.21\%                       & 0.6835                       \\
D3                                            & 97.89\%                    & 84.36\%                       & 0.9062                        & 70.46\%                    & 62.78\%                       & 0.6640                       \\
D4                                            & 98.90\%                    & 97.83\%                       & 0.9836                        & 73.63\%                    & 72.83\%                       & 0.7323                       \\
D5                                            & 99.80\%                    & 88.49\%                       & 0.9381                        & 68.02\%                    & 77.19\%                       & 0.7232                       \\
D6                                            & 98.31\%                    & 93.19\%                       & 0.9568                        & 71.46\%                    & 86.01\%                       & 0.7806                       \\
D7                                            & 99.15\%                    & 97.48\%                       & 0.9831                        & 65.81\%                    & 67.46\%                       & 0.6662                       \\
D8                                            & 97.14\%                    & 97.14\%                       & 0.9714                        & 71.43\%                    & 86.21\%                       & 0.7813                       \\
D9                                            & 99.74\%                    & 85.16\%                       & 0.9188                        & 74.52\%                    & 63.17\%                       & 0.6838                       \\
D10                                           & 98.48\%                    & 87.80\%                       & 0.9283                        & 75.10\%                    & 64.02\%                       & 0.6912                       \\ \midrule
\textbf{Avg.}                                 & \textbf{}                  & \textbf{}                     & \textbf{0.9434}               & \textbf{}                  & \textbf{}                     & \textbf{0.7137}              \\ \bottomrule
\end{tabular}
\end{table}

We analyze the reason as follows. The cost of comparing each pair of logic gates in the task of learning functionality is extremely high, which is proportional to the square of the circuit size. We limit the dataset and train the model to learn functional similarity only among pairs with similar logic probability, which is a necessary condition for functional equivalence. Therefore, without the staged multi-stage strategy, be effectively supervised with the simplified dataset, leading to poor performance in learning functionality. 
As shown in Table~\ref{TAB:ABS:Loss}, the differences between the two models in the loss values for predicting logic probability ($L_{prob}$) and identifying reconvergence structures ($L_{rc}$) are not significant, indicating that they perform similarly in these two tasks. However, compared to the w/o multi-stage model, the w/ multi-stage model performs better in learning functionality with $L_{func} = 0.0594$, which is $51.47\%$ smaller than that of w/o multi-stage model. 
However, the w/ multi-stage model outperforms the model w/o multi-stage in learning functionality task with a significantly lower $L_{func}$ value of $0.0594$, which is $51.47\%$ smaller than that of the latter.

\begin{table}[!t]
\centering
\caption{Loss Comparison between w/ multi-stage and w/o multi-stage} \label{TAB:ABS:Loss}
\begin{tabular}{@{}llll@{}}
\toprule
           & \textit{w/o multi-stage} & \textit{w/ multi-stage} & \textbf{Reduction}   \\ \midrule
$L_{prob}$ & 0.0205         & 0.0207          & \textbf{-0.98\%}  \\
$L_{rc}$   & 0.1186         & 0.1115          & \textbf{5.99\%}   \\
$L_{func}$ & 0.1224         & 0.0594          & \textbf{51.47\%} \\ \bottomrule
\end{tabular}
\vspace{-10pt}
\end{table}


\section{Downstream Tasks} \label{Sec:Task}
In this section, we combine our DeepGate2 with the open-source EDA tools and apply our model to practical EDA tasks: logic synthesis and Boolean satisfiability (SAT) solving. The logic synthesis tools aim to identify logic equivalence gates as quickly as possible. In Section~\ref{Sec:Task:Sweep}, our proposed functionality-aware circuit learning model provides guidance to the logic synthesis tool about the logic similarity. Additionally, in Section~\ref{Sec:Task:SAT}, we apply the learnt functional similarity in SAT solving, where the variables with dissimilar functionality are assigned the same decision value. This approach efficiently shrinks the search space by enabling solvers to encounter more constraints.

\subsection{Logic Synthesis} \label{Sec:Task:Sweep}
\begin{figure}[!t]
    \centering
    \includegraphics[width=1.0\linewidth]{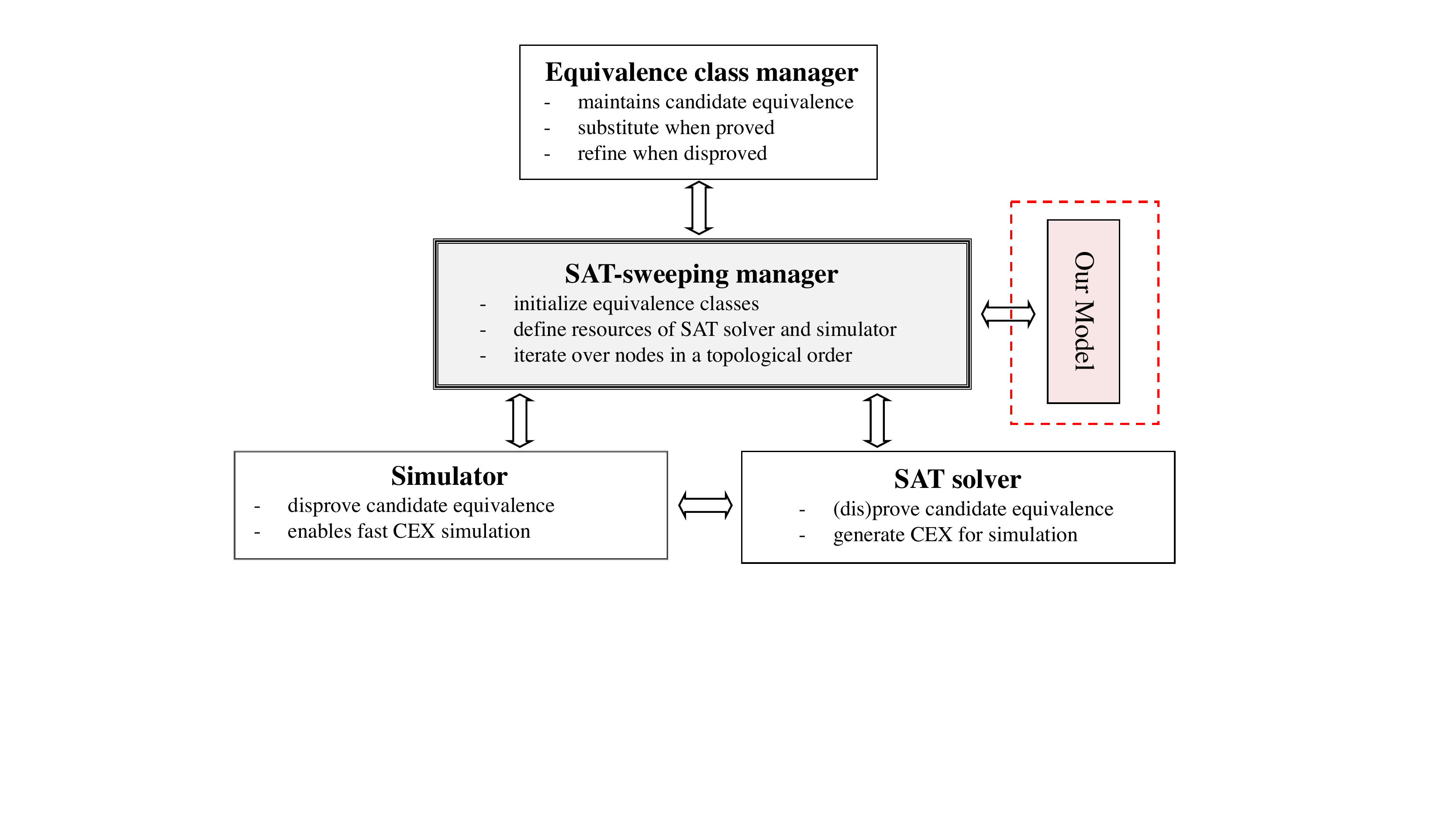}
    \vspace{-10pt}
    \caption{The proposed SAT-sweeping ecosystem.}
    \vspace{-15pt}
    \label{fig:SATsweep}
\end{figure}

This subsection shows the effectiveness of our proposed functionality-aware circuit learning framework in SAT-sweeping~\cite{kuehlmann2002robust}, a common technique of logic synthesis. Fig.~\ref{fig:SATsweep} illustrates the components of a typical ecosystem for SAT-sweeping engine (also called SAT sweeper), where including \emph{equivalence class (EC) manager}, \emph{SAT-sweeping manager}, \emph{simulator}, and \emph{SAT solver}. All computations are coordinated by the SAT-sweeping manager~\cite{mishchenko2018integrating}. The SAT sweeper starts by computing candidate ECs using several rounds of initial simulation and storing ECs into EC manager. In the next step, the SAT-sweeping manager selects two gates within an EC and then calls the SAT solver to check whether they are equivalent. If so, the EC manager merges these two gates. Otherwise, SAT solver will return a satisfiable assignment as a counterexample for incremental simulation to refine the candidate ECs.

To the best of our knowledge, most SAT-sweeping managers select EC only based on the circuit structure, without efficient heuristic strategy considering the functionality of candidate gates. We will introduce the functional information into SAT-sweeping manager to further improve efficiency. 

\subsubsection{Experiment Settings} 

We combine our DeepGate2 into SAT sweeper to guide EC selection. To be specific, the updated manager sorts all candidate equivalence classes by computing the cosine similarity of their embeddings. Unlike traditional SAT sweepers, our proposed SAT sweeper does not need to validate the equivalence of all candidate ECs in one pass. Instead, node pairs with higher similarity have high priority for SAT solver calls. If the gate pair is formally proved to be equivalent, these two gate are merged. Otherwise, the generated counterexample should contain more conflicts than the baseline method, resulting in better efficiency for refining candidate ECs.

Our model is equipped into ABC~\cite{abc} as a plug-in and integrated into the SAT sweeper `\texttt{\&fraig}'~\cite{mishchenko2005fraigs}, which is one of the most efficient and scalable SAT sweeper publicly available at this time. The AIGs derived by merging the resulting equivalence nodes are verified by `\texttt{\&cec}' command in ABC to ensure functional correctness. All experiments are performed on a 2.40 GHz Intel(R) Xeon(R) Silver 4210R CPU with 64GB of main memory. A single core and less than 1GB was used for any test case considered in this subsection. The proposed SAT sweeper (named as \emph{Our}) is compared against the original engine, \emph{\&fraig}.

\subsubsection{Results}

We validate the performance of our SAT sweeper with $6$ industrial circuits. As shown in Table~\ref{TAB:SATsweep}, section ``Statistics'' lists the number of PI and PO (PI/PO), logic levels (Lev) and internal AND-nodes in the original AIG (And). To ensure the fairness of comparison, the circuits after sweeping should have the same size. 
Section ``SAT calls'' lists the number of satisfiable SAT calls, performed by the solver employed in each engine. The data shows that our proposed engine decreases the number of satisfiable SAT calls, that explains why it has better results, since more resource are used to prove equivalence gates. In addition, section ``Total runtime'' compares the runtime and section ``Red.'' shows the runtime reduction from \&fraig to Our. 

The number of ``SAT calls'' can get an average reduction of $53.37\%$ ($95.88\%$ maximum) through the integration of our DeepGate2 model. As for ``Total runtime'', this experiment shows that the DeepGate2-based SAT sweeper outperforms state-of-the-art engines, while reducing the average runtime by $49.46\%$ ($57.77\%$ maximum). Thus, the sweeper formally verifies the equivalence of functionally similar gates with the guidance from DeepGate2, thereby reducing the number of invalid SAT calls and improving efficiency of SAT sweeping. Take C1 as a representative example, the baseline \&fraig selects gates in EC for formal verification without considering their behaviour, thus, many solver calls return satisfiable results, and few gates can be merged. However, with the guidance of DeepGate2, the sweeper can prioritize the selection of gates with similar behavior, resulting in a significant reduction of $95.88\%$ in SAT calls and $57.77\%$ in runtime.

\begin{table}[!t]
\centering
\caption{Comparing the number of SAT calls and the runtime of the SAT sweepers} \label{TAB:SATsweep}
\setlength\tabcolsep{2.0pt}
\begin{tabular}{@{}l|lll|lll|lll@{}}
\toprule
\multicolumn{1}{c|}{\multirow{2}{*}{Circuit}} & \multicolumn{3}{c|}{Statistic}    & \multicolumn{3}{c|}{SAT calls}           & \multicolumn{3}{c}{Total Runtime (s)}    \\
\multicolumn{1}{c|}{}                         & PI/PO     & Lev       & And       & \&fraig   & Our       & Red.             & \&fraig   & Our       & Red.             \\ \midrule
C1                                            & 128/128   & 4,372     & 57,247    & 13,826    & 570       & 95.88\%          & 7.46      & 3.15      & 57.77\%          \\
C2                                            & 24/25     & 225       & 5,416     & 100       & 74        & 26.00\%          & 5.43      & 3.11      & 42.73\%          \\
C3                                            & 22/1      & 29        & 703       & 4         & 1         & 75.00\%          & 19.07     & 9.10      & 52.28\%          \\
C4                                            & 114/1     & 91        & 19,354    & 20        & 12        & 40.00\%          & 6.49      & 4.04      & 37.75\%          \\
C5                                            & 126/1     & 83        & 20,971    & 6         & 4         & 33.33\%          & 0.48      & 0.21      & 56.25\%          \\
C6                                            & 96/1      & 79        & 14,389    & 10        & 5         & 50.00\%          & 0.30      & 0.15      & 50.00\%          \\ \midrule
\textbf{Avg.}                                 & \textbf{} & \textbf{} & \textbf{} & \textbf{} & \textbf{} & \textbf{53.37\%} & \textbf{} & \textbf{} & \textbf{49.46\%} \\ \bottomrule
\end{tabular}
\vspace{-10pt}
\end{table}

\subsection{Boolean Satisfiability Solving} \label{Sec:Task:SAT}

Boolean satisfiability (SAT) solving is a long-standing and fundamental NP-complete problem with applications in many areas, especially in electronic design automation (EDA)~\cite{goldberg2001using, mcmillan2003interpolation, yang2004trangen}. The existing SAT solvers are designed to incorporate efficient heuristics~\cite{lu2003circuit, audemard2009glucose, audemard2018glucose} to expedite the solving process. For instance, \cite{lu2003circuit} proposes to utilize the correlation of logic gate functionality to enforce variable decision for solving circuit-based SAT instances. Although the solution achieves remarkable speedup over SAT solvers, it still relies on the time-consuming logic simulation to obtain the functionality. Based on~\cite{lu2003circuit}, we demonstrate how the DeepGate2 models functional correlation efficiently and accelerates SAT solving. 

\subsubsection{Experiment Settings}
We integrate our DeepGate2 into a modern SAT solver, CaDiCal~\cite{queue2019cadical} to solve the instances from logic equivalence checking (LEC) task. Firstly, we obtain gate-level embeddings of the original circuit and predict the pairwise functional similarity between these gates. Given the one-to-one mapping~\cite{tseitin1983complexity} between circuits and conjunctive normal form (a problem format required by SAT solvers), we can easily transfer the gate functional similarity to variable behavioral similarity. If two logic gates have similar representations (and therefore similar functionality), their corresponding variables should be correlated and grouped together during the variable decision process. 

Secondly, we incorporate the learnt knowledge into the SAT solver. As shown in Algorithm~\ref{Boolean-Satisfiability-Solving}, when the current variable $s$ is assigned a value $v$, we identify all unassigned variables $s^{\prime}$ in the set $\mathcal{S}$ that contains correlated variables with $s$. As modern SAT solvers reduce searching space by detecting conflicts as much as possible~\cite{marques2021conflict}, we assign the reverse value $\Bar{v}$ to $s^{\prime}$ to promptly cause conflict for joint decision. Besides, the threshold $\delta$ in Algorithm~\ref{Boolean-Satisfiability-Solving} is set to $1e-5$.  

Thirdly, to evaluate the efficacy of our model in accelerating SAT solving, we compare the aforementioned hybrid solver (labeled as Our) with original CaDiCal~\cite{queue2019cadical} (labeled as Baseline) on $5$ industrial instances. All experiments are conducted with a single 2.40GHz Intel(R) Xeon(R) E5-2640 v4 CPU. 

\begin{algorithm}[htbp]
    \caption{Variable Decision Function with DeepGate2}
    \label{Boolean-Satisfiability-Solving}
    Current variable $s$ just being assigned a value $v$; Set $\mathcal{S}$ containing the correlated variables with $s$; \\
    Function $\text{Sim}(s_i,s_j)$ to calculate the behaviour similarity of two variables; \\
    $\delta$ is the threshold for the joint decision \\
    Function $V(s)$ to get the assigned value of variable $s$; \\
    Function $\text{Decision}(s, v)$ to assign the decision value $v$ to the current decision variable $s$. 
    \begin{algorithmic}[1]
    \STATE \text{Decision}($s$, $v$)
    \FOR{$s^{\prime}$ in $\mathcal{S}$}
    \IF{$s^{\prime} \neq s$ and $V(s^{\prime}) = \text{None}$} 
    \IF{$\text{Sim}(s^{\prime}, s) > 1 - \delta$}
    \STATE \text{Decision}($s^{\prime}$, $\Bar{v}$)
    \ENDIF
    \ENDIF
    \ENDFOR
    \end{algorithmic}
\end{algorithm}

\subsubsection{Results}
\begin{table}[!t]
\centering
\renewcommand\tabcolsep{5.0pt}
\caption{Comparing the Runtime between Baseline and Our Solvers} \label{TAB:SATSolving}
\begin{tabular}{@{}ll|l|lll|l@{}} \toprule
\multicolumn{1}{c}{\multirow{2}{*}{Instance}} & \multicolumn{1}{c|}{\multirow{2}{*}{Size}} & Baseline(s)  & \multicolumn{3}{c|}{Our(s)} & \multicolumn{1}{c}{\multirow{2}{*}{Reduction}} \\
\multicolumn{1}{c}{}                          & \multicolumn{1}{c|}{}                      & & Model      & Solver    & Overall   & \multicolumn{1}{c}{}                           \\ \midrule
I1                                            & 17,495                                     & 88.01     & 1.77       & 30.25     & 32.02     & 63.62\%                                        \\
I2                                            & 21,952                                     & 29.36     & 2.85       & 6.01      & 8.86      & 69.82\%                                        \\
I3                                            & 23,810                                     & 61.24     & 3.25       & 32.88     & 36.13     & 41.00\%                                        \\
I4                                            & 27,606                                     & 158.04    & 4.36       & 137.77    & 142.13    & 10.07\%                                        \\
I5                                            & 28,672                                     & 89.89     & 4.78       & 70.95     & 75.73     & 15.75\%                                        \\ \midrule
\textbf{Avg.}                                 & \textbf{}                                  & \textbf{} & \textbf{}  & \textbf{} & \textbf{} & \textbf{40.05\%} \\     \bottomrule                        
\end{tabular}
\vspace{-10pt}
\end{table}

The runtime comparison between Baseline and Our are listed in Table~\ref{TAB:SATSolving}. To ensure a fair comparison, we aggregate the DeepGate2 model inference time (Model) and SAT solver runtime (Solver) as the Overall runtime. We have the following observations.
First, our method achieves a substantial reduction in total runtime for all test cases, with an average runtime reduction of $40.05\%$. Take I1 as an example, the plain solver requires $88.01$s to solve the problem, but by combining with our model, the new solver produces results in only $32.02$s, reducing runtime by $63.62\%$. 
Second, our model only takes a few seconds to obtain embeddings, occupying less than $10\%$ of overall runtime on average. It should be noted that our DeepGate2 is able to infer within polynomial time that is only proportional to the size of instance. 
Third, while the two largest instances I4 and I5 show less reduction than the others, it does not necessarily mean that our model is unable to generalize to larger instances. As evidenced by the results for I2, an instance with a similar size to I4 and I5 also demonstrates a significant reduction. The reduction caused by our model should be determined by the characteristics of instance. 
In summary, our model is effective in speeding up downstream SAT solving.

\section{Conclusion}
\label{Sec:Conclusion}
This paper introduces DeepGate2, a novel functionally-aware framework for circuit representation learning. Our approach leverages the pairwise truth table differences of logic gates as a supervisory signal, providing rich functionality supervision and proving scalable for large circuits. Moreover, DeepGate2 differentiates and concurrently updates structural and functional embeddings in two dedicated flows, acquiring comprehensive representations through a single round of GNN forward message-passing. In comparison to its predecessor, DeepGate2 demonstrates enhanced performance in logic probability prediction and logic equivalent gate identification, while simultaneously improving model efficiency tenfold. The applications of DeepGate2 onto multiple downstream tasks further demonstrate its effectiveness and potential utility in the EDA field.

\clearpage
\balance
\bibliographystyle{IEEEtran}

\begin{thebibliography}{10}
\providecommand{\url}[1]{#1}
\csname url@samestyle\endcsname
\providecommand{\newblock}{\relax}
\providecommand{\bibinfo}[2]{#2}
\providecommand{\BIBentrySTDinterwordspacing}{\spaceskip=0pt\relax}
\providecommand{\BIBentryALTinterwordstretchfactor}{4}
\providecommand{\BIBentryALTinterwordspacing}{\spaceskip=\fontdimen2\font plus
\BIBentryALTinterwordstretchfactor\fontdimen3\font minus
  \fontdimen4\font\relax}
\providecommand{\BIBforeignlanguage}[2]{{%
\expandafter\ifx\csname l@#1\endcsname\relax
\typeout{** WARNING: IEEEtran.bst: No hyphenation pattern has been}%
\typeout{** loaded for the language `#1'. Using the pattern for}%
\typeout{** the default language instead.}%
\else
\language=\csname l@#1\endcsname
\fi
#2}}
\providecommand{\BIBdecl}{\relax}
\BIBdecl

\bibitem{chen2020pros}
J.~Chen, J.~Kuang, G.~Zhao, D.~J.-H. Huang, and E.~F. Young, ``Pros: A plug-in
  for routability optimization applied in the state-of-the-art commercial eda
  tool using deep learning,'' in \emph{Proceedings of the 39th International
  Conference on Computer-Aided Design}, 2020, pp. 1--8.

\bibitem{mirhoseini2021graph}
A.~Mirhoseini, A.~Goldie, M.~Yazgan, J.~W. Jiang, E.~Songhori, S.~Wang, Y.-J.
  Lee, E.~Johnson, O.~Pathak, A.~Nazi \emph{et~al.}, ``A graph placement
  methodology for fast chip design,'' \emph{Nature}, vol. 594, no. 7862, pp.
  207--212, 2021.

\bibitem{neto2019lsoracle}
W.~L. Neto, M.~Austin, S.~Temple, L.~Amaru, X.~Tang, and P.-E. Gaillardon,
  ``Lsoracle: A logic synthesis framework driven by artificial intelligence,''
  in \emph{2019 IEEE/ACM International Conference on Computer-Aided Design
  (ICCAD)}.\hskip 1em plus 0.5em minus 0.4em\relax IEEE, 2019, pp. 1--6.

\bibitem{haaswijk2018deep}
W.~Haaswijk, E.~Collins, B.~Seguin, M.~Soeken, F.~Kaplan, S.~S{\"u}sstrunk, and
  G.~De~Micheli, ``Deep learning for logic optimization algorithms,'' in
  \emph{2018 IEEE International Symposium on Circuits and Systems
  (ISCAS)}.\hskip 1em plus 0.5em minus 0.4em\relax IEEE, 2018, pp. 1--4.

\bibitem{shi2022deeptpi}
Z.~Shi, M.~Li, S.~Khan, L.~Wang, N.~Wang, Y.~Huang, and Q.~Xu, ``Deeptpi: Test
  point insertion with deep reinforcement learning,'' in \emph{2022 IEEE
  International Test Conference (ITC)}.\hskip 1em plus 0.5em minus 0.4em\relax
  IEEE, 2022, pp. 194--203.

\bibitem{huang2022neural}
J.~Huang, H.-L. Zhen, N.~Wang, H.~Mao, M.~Yuan, and Y.~Huang, ``Neural fault
  analysis for sat-based atpg,'' in \emph{2022 IEEE International Test
  Conference (ITC)}.\hskip 1em plus 0.5em minus 0.4em\relax IEEE, 2022, pp.
  36--45.

\bibitem{li2022deepgate}
M.~Li, S.~Khan, Z.~Shi, N.~Wang, H.~Yu, and Q.~Xu, ``Deepgate: Learning neural
  representations of logic gates,'' in \emph{Proceedings of the 59th ACM/IEEE
  Design Automation Conference}, 2022, pp. 667--672.

\bibitem{wang2022functionality}
Z.~Wang, C.~Bai, Z.~He, G.~Zhang, Q.~Xu, T.-Y. Ho, B.~Yu, and Y.~Huang,
  ``Functionality matters in netlist representation learning,'' in
  \emph{Proceedings of the 59th ACM/IEEE Design Automation Conference}, 2022,
  pp. 61--66.

\bibitem{zhu2022tag}
K.~Zhu, H.~Chen, W.~J. Turner, G.~F. Kokai, P.-H. Wei, D.~Z. Pan, and H.~Ren,
  ``Tag: Learning circuit spatial embedding from layouts,'' in
  \emph{Proceedings of the 41st IEEE/ACM International Conference on
  Computer-Aided Design}, 2022, pp. 1--9.

\bibitem{lai2022maskplace}
Y.~Lai, Y.~Mu, and P.~Luo, ``Maskplace: Fast chip placement via reinforced
  visual representation learning,'' \emph{arXiv preprint arXiv:2211.13382},
  2022.

\bibitem{fayyazi2019deep}
A.~Fayyazi, S.~Shababi, P.~Nuzzo, S.~Nazarian, and M.~Pedram, ``Deep
  learning-based circuit recognition using sparse mapping and level-dependent
  decaying sum circuit representations,'' in \emph{2019 Design, Automation \&
  Test in Europe Conference \& Exhibition (DATE)}.\hskip 1em plus 0.5em minus
  0.4em\relax IEEE, 2019, pp. 638--641.

\bibitem{he2021graph}
Z.~He, Z.~Wang, C.~Bail, H.~Yang, and B.~Yu, ``Graph learning-based arithmetic
  block identification,'' in \emph{2021 IEEE/ACM International Conference On
  Computer Aided Design (ICCAD)}.\hskip 1em plus 0.5em minus 0.4em\relax IEEE,
  2021, pp. 1--8.

\bibitem{li2022deepsat}
M.~Li, Z.~Shi, Q.~Lai, S.~Khan, and Q.~Xu, ``Deepsat: An eda-driven learning
  framework for sat,'' \emph{arXiv preprint arXiv:2205.13745}, 2022.

\bibitem{mishchenko2005fraigs}
A.~Mishchenko, S.~Chatterjee, R.~Jiang, and R.~K. Brayton, ``Fraigs: A unifying
  representation for logic synthesis and verification,'' ERL Technical Report,
  Tech. Rep., 2005.

\bibitem{queue2019cadical}
S.~D. QUEUE, ``Cadical at the sat race 2019,'' \emph{SAT RACE 2019}, p.~8,
  2019.

\bibitem{brown2020language}
T.~Brown, B.~Mann, N.~Ryder, M.~Subbiah, J.~D. Kaplan, P.~Dhariwal,
  A.~Neelakantan, P.~Shyam, G.~Sastry, A.~Askell \emph{et~al.}, ``Language
  models are few-shot learners,'' \emph{Advances in neural information
  processing systems}, vol.~33, pp. 1877--1901, 2020.

\bibitem{devlin2018bert}
J.~Devlin, M.-W. Chang, K.~Lee, and K.~Toutanova, ``Bert: Pre-training of deep
  bidirectional transformers for language understanding,'' \emph{arXiv preprint
  arXiv:1810.04805}, 2018.

\bibitem{wu2018unsupervised}
Z.~Wu, Y.~Xiong, S.~X. Yu, and D.~Lin, ``Unsupervised feature learning via
  non-parametric instance discrimination,'' in \emph{Proceedings of the IEEE
  conference on computer vision and pattern recognition}, 2018, pp. 3733--3742.

\bibitem{hoffer2015deep}
E.~Hoffer and N.~Ailon, ``Deep metric learning using triplet network,'' in
  \emph{Similarity-Based Pattern Recognition: Third International Workshop,
  SIMBAD 2015, Copenhagen, Denmark, October 12-14, 2015. Proceedings 3}.\hskip
  1em plus 0.5em minus 0.4em\relax Springer, 2015, pp. 84--92.

\bibitem{vaswani2017attention}
A.~Vaswani, N.~Shazeer, N.~Parmar, J.~Uszkoreit, L.~Jones, A.~N. Gomez,
  {\L}.~Kaiser, and I.~Polosukhin, ``Attention is all you need,''
  \emph{Advances in neural information processing systems}, vol.~30, 2017.

\bibitem{bengio2009curriculum}
Y.~Bengio, J.~Louradour, R.~Collobert, and J.~Weston, ``Curriculum learning,''
  in \emph{Proceedings of the 26th annual international conference on machine
  learning}, 2009, pp. 41--48.

\bibitem{ruder2017overview}
S.~Ruder, ``An overview of multi-task learning in deep neural networks,''
  \emph{arXiv preprint arXiv:1706.05098}, 2017.

\bibitem{ITC99}
S.~Davidson, ``Characteristics of the itc’99 benchmark circuits,'' in
  \emph{ITSW}, 1999.

\bibitem{albrecht2005iwls}
C.~Albrecht, ``Iwls 2005 benchmarks,'' in \emph{IWLS}, 2005.

\bibitem{EPFLBenchmarks}
L.~Amar{\'u}, P.-E. Gaillardon, and G.~De~Micheli, ``The epfl combinational
  benchmark suite,'' in \emph{IWLS}, no. CONF, 2015.

\bibitem{takeda2008opencore}
O.~Team, ``Opencores,'' \url{https://opencores.org/}.

\bibitem{kingma2014adam}
D.~P. Kingma and J.~Ba, ``Adam: A method for stochastic optimization,''
  \emph{arXiv preprint arXiv:1412.6980}, 2014.

\bibitem{kuehlmann2002robust}
A.~Kuehlmann, V.~Paruthi, F.~Krohm, and M.~K. Ganai, ``Robust boolean reasoning
  for equivalence checking and functional property verification,'' \emph{IEEE
  Transactions on Computer-Aided Design of Integrated Circuits and Systems},
  vol.~21, no.~12, pp. 1377--1394, 2002.

\bibitem{mishchenko2018integrating}
A.~Mishchenko and R.~Brayton, ``Integrating an aig package, simulator, and sat
  solver,'' in \emph{International Workshop on Logic and Synthesis (IWLS)},
  2018, pp. 11--16.

\bibitem{abc}
B.~L. Synthesis and V.~Group, ``Abc: A system for sequential synthesis and
  verification.'' \emph{http://www-cad.eecs.berkeley.edu/\~alanmi/abc}, 2023.

\bibitem{goldberg2001using}
E.~I. Goldberg, M.~R. Prasad, and R.~K. Brayton, ``Using sat for combinational
  equivalence checking,'' in \emph{Proceedings Design, Automation and Test in
  Europe. Conference and Exhibition 2001}.\hskip 1em plus 0.5em minus
  0.4em\relax IEEE, 2001, pp. 114--121.

\bibitem{mcmillan2003interpolation}
K.~L. McMillan, ``Interpolation and sat-based model checking,'' in
  \emph{Computer Aided Verification: 15th International Conference, CAV 2003,
  Boulder, CO, USA, July 8-12, 2003. Proceedings 15}.\hskip 1em plus 0.5em
  minus 0.4em\relax Springer, 2003, pp. 1--13.

\bibitem{yang2004trangen}
K.~Yang, K.-T. Cheng, and L.-C. Wang, ``Trangen: A sat-based atpg for
  path-oriented transition faults,'' in \emph{ASP-DAC 2004: Asia and South
  Pacific Design Automation Conference 2004 (IEEE Cat. No. 04EX753)}.\hskip 1em
  plus 0.5em minus 0.4em\relax IEEE, 2004, pp. 92--97.

\bibitem{lu2003circuit}
F.~Lu, L.-C. Wang, K.-T. Cheng, and R.-Y. Huang, ``A circuit sat solver with
  signal correlation guided learning,'' in \emph{2003 Design, Automation and
  Test in Europe Conference and Exhibition}.\hskip 1em plus 0.5em minus
  0.4em\relax IEEE, 2003, pp. 892--897.

\bibitem{audemard2009glucose}
G.~Audemard and L.~Simon, ``Glucose: a solver that predicts learnt clauses
  quality,'' \emph{SAT Competition}, pp. 7--8, 2009.

\bibitem{audemard2018glucose}
------, ``On the glucose sat solver,'' \emph{International Journal on
  Artificial Intelligence Tools}, vol.~27, no.~01, p. 1840001, 2018.

\bibitem{tseitin1983complexity}
G.~S. Tseitin, ``On the complexity of derivation in propositional calculus,''
  \emph{Automation of reasoning: 2: Classical papers on computational logic
  1967--1970}, pp. 466--483, 1983.

\bibitem{marques2021conflict}
J.~Marques-Silva, I.~Lynce, and S.~Malik, ``Conflict-driven clause learning sat
  solvers,'' in \emph{Handbook of satisfiability}.\hskip 1em plus 0.5em minus
  0.4em\relax IOS press, 2021, pp. 133--182.

\end{thebibliography}

\end{document}